\documentclass[10pt,twocolumn,letterpaper]{article}
\usepackage[accsupp]{axessibility} 
 \usepackage{cvpr}              

\usepackage[dvipsnames]{xcolor}

\usepackage{bm}
\usepackage{amsmath}

\usepackage{tikz, pgfplots}
\usepackage{pgfplotstable}
\usepgfplotslibrary{colorbrewer}
\usetikzlibrary{pgfplots.statistics, pgfplots.colorbrewer} 
\usepackage{float} 
\usepackage{graphicx}
\usepackage{caption}
\usepackage{multirow}
\usepackage{siunitx}
\usepackage{courier}
\usepgfplotslibrary{fillbetween}
\usetikzlibrary{fadings}
\usepackage{subcaption}
\usetikzlibrary{arrows, calc,intersections, shapes}
\usetikzlibrary{arrows.meta}
\usetikzlibrary{decorations.pathmorphing}
\usetikzlibrary{decorations.text}
\usetikzlibrary{decorations.pathreplacing}
\usetikzlibrary{hobby,decorations.markings}
\usetikzlibrary{positioning}
\usetikzlibrary{calc}
\usepackage{tabularx}
\usetikzlibrary{shapes.multipart}
\newcolumntype{C}{>{\centering\arraybackslash}X}

\pgfmathdeclarefunction{gauss}{3}{%
	\pgfmathparse{1/(#3*sqrt(2*pi))*exp(-((#1-#2)^2)/(2*#3^2))}%
}
\usetikzlibrary{patterns}
\usepackage[switch]{lineno}  
\def\BibTeX{{\rm B\kern-.05em{\sc i\kern-.025em b}\kern-.08em
		T\kern-.1667em\lower.7ex\hbox{E}\kern-.125emX}}

\definecolor{blue0}{HTML}{0077ff}
\definecolor{green0}{HTML}{00e89d}
\definecolor{orange0}{HTML}{ffb067}

\definecolor{cvprblue}{rgb}{0.21,0.49,0.74}
\usepackage[pagebackref,breaklinks,colorlinks,citecolor=cvprblue]{hyperref}

\def\paperID{8} 
\def\confName{CVPR}
\def\confYear{2024}

\title{Reliable Trajectory Prediction and Uncertainty Quantification\\with Conditioned Diffusion Models}

\author{Marion Neumeier\textsuperscript{1}\quad Sebastian Dorn\textsuperscript{2,3}\quad  Michael Botsch\textsuperscript{1} \quad Wolfgang Utschick\textsuperscript{4}\\
\textsuperscript{1}Technische Hochschule Ingolstadt, \quad \quad \textsuperscript{2}Audi AG,\\\textsuperscript{3}Technische Hochschule Augsburg, \quad \quad \textsuperscript{4}Technische Universität München\\
{\tt\small \{marion.neumeier, michael.botsch\}@thi.de, sebastian.dorn@tha.de, utschick@tum.de}
}
\begin{document}
\maketitle
\setlength{\abovedisplayskip}{4.5pt}
\setlength{\belowdisplayskip}{4.5pt}
\begin{abstract}
This work introduces the conditioned Vehicle Motion Diffusion (cVMD) model, a novel network architecture for highway trajectory prediction using diffusion models. The proposed model ensures the drivability of the predicted trajectory by integrating non-holonomic motion constraints and physical constraints into the generative prediction module. Central to the architecture of cVMD is its capacity to perform uncertainty quantification, a feature that is crucial in safety-critical applications. By integrating the quantified uncertainty into the prediction process, the cVMD's trajectory prediction performance is improved considerably. The model’s performance was evaluated using the publicly available highD dataset. Experiments show that the proposed architecture achieves competitive trajectory prediction accuracy compared to state-of-the-art models, while providing guaranteed drivable trajectories and uncertainty quantification.
\end{abstract}
    
\section{Introduction}
\label{sec:intro}
Vehicle trajectory prediction is a fundamental challenge in the automotive domain \cite{CVPR2022Bahari, bharilya2023machine}. Due to the highly interactive nature of traffic scenarios, model-based approaches are generally not able to capture or represent the underlying complexity and the variety of traffic situations. Many prominent approaches for predicting trajectories in cooperative traffic scenarios apply data-driven algorithms, \mbox{\eg \cite{AlexandreAlahi.2016, Deo.2018}}. While these approaches are capable of successfully modelling driving behaviour, the feasibility and drivability of the predicted trajectories are not guaranteed. 
Vehicles are non-holonomic systems with restricted movement capabilities, such as the coupling between forward and sideway motion.
Most machine learning (ML)-based vehicle motion prediction models do not account for non-holonomic and general physical constraints \cite{9756903}. As a result, there is no guarantee that predictions of ML models are realistic or consistent with the general constraints of motion \cite{Deo.2018, Messaoud.2021}. Another shortcoming of data-driven regression models is that they typically lack the ability to quantify the uncertainty in their predictions \cite{LAI2022249, GAWLI2022, ayhan2018testtime, NEURIPS2020_322f6246}. The models typically provide point estimates that provide the most likely prediction, but do not account for uncertainty in future trajectories. 
In safety-critical applications, however, it is crucial to have knowledge of the uncertainties associated with trajectory predictions. This enables intelligent systems to make informed decisions and mitigate potential risks \cite{SUK2024317}.
The aim of this work is to address these limitations by introducing the conditioned Vehicle Motion Diffusion Model (cVMD). cVMD is composed of a classifier-free guided diffusion-based probabilistic model considering the non-holonomic kinematic constraints of vehicles. The trajectory prediction task is regarded as a reverse diffusion process, conditional on an interactive highway traffic scenario. To understand the traffic scenario context, cVMD integrates a Vector Quantized Variational Autoencoder (VQ-VAE) as illustrated in Fig.~\ref{fig:vmd_architecture}. VQ-VAE effectively discretizes the infinite traffic scenario constellations into distinct representative contexts. 
The cVMD architecture inherently allows for the quantification of the uncertainty in the model's predictions.
This uncertainty quantification is used to make uncertainty-adaptive trajectory predictions.
The main contributions are as follows:
\begin{itemize} 
	\item Introduction of the cVMD architecture for the prediction of guaranteed drivable trajectories.
	\item Proposal of a method to quantify prediction uncertainty and to integrate it into trajectory prediction.
	\item Leveraging the model's generative capabilities to represent real-world scenario stochasticity.
	\item Evaluation of prediction performance on publicly available highD dataset.
\end{itemize}

\section{Related Work}
\label{sec:relatedwork}
Modeling the complex interactions in traffic scenarios and their impact on individual driving behaviors poses a significant challenge. Consequently, many studies on trajectory predictions rely on data-driven methods, \eg \cite{AlexandreAlahi.2016, Deo.2018, Neumeier.2021, Seff.2023}. Recently, there has been a growing emphasis on graph-based approaches \cite{DefuCao.2021, Neumeier.2022, HaoZhou.2021, Ding.2021} as they allow to directly model relations and inter-dependencies. However, it has been shown that graph-based approaches come with limitations \cite{Neumeier.2023, NT.2019}. 
As a result of the remarkable achievements in the areas of computer vision \cite{Ramesh.2023, Saharia.2022} and natural language processing \cite{Gong.2022, Li.2022}, diffusion models are gaining popularity in a wide range of fields, including the automotive domain. 
For example, in the work of \cite{Zhong.2022}, the authors propose a diffusion model for controllable traffic scenario generation. The sampling process of the diffusion model is guided by specific scenario conditions, which allow for the generation of diverse yet controllable scenarios. Furthermore, Provnost \etal~\cite{Pronovost.2023} introduce a latent diffusion model \cite{Ramesh.2023} that utilizes map data to generate realistic driving scenes. Similarly, the authors of \cite{Balasubramanian.2023} use a conditional latent diffusion model with a temporal constraint for scene prediction. The focus of their work is on scenario generation rather than trajectory prediction. In their work, the authors also reconstruct the map data. This is an additional task on top of the scene prediction, introducing an avoidable level of complexity.
Chen \etal~\cite{Chen.2023} introduce EquiDiff, a deep generative diffusion model for predicting vehicle trajectories based on historical scenario information. The historical scenario information is embedded using Gated Recurrent Unit~\cite{GRU} and Graph Attention Network~\cite{GraphGAT} and subsequently provided as contextual information for the trajectory generation process. By providing the contextual information the generated trajectory prediction is additionally conditioned on the observed scenario information. The intended effect of additional conditioning is to decrease generative diversity of the trajectory prediction while increasing the likelihood of a trajectory close to the ground truth future trajectory. It is worth noting, however, that
the conditioning approach of the diffusion process applied by the authors differs from that proposed by Ho \etal~\cite{Ho.2022}. 
Based on the analysis of the existing literature, several works have used diffusion models to predict vehicle motion. However, a common limitation of these methods is the lack of guaranteed trajectory feasibility.
The authors of \cite{Tevet.2023} propose the Human Motion Diffusion Model (MDM), a classifier-free diffusion-based model for the generation of realistic human motion. MDM predicts the samples rather than the noise for each diffusion step, allowing the addition of geometric losses such as foot contact to improve human motion synthesis. The optimization of the Transformer-based \cite{Vaswani.2017} MDM ensures that the generative process aligns with both the general abilities of humans and the principles of physics. Despite the extensive research, none of the above studies have incorporated uncertainty quantification of their motion predictions. 
\section{Preliminaries}
\subsection{Denoising Diffusion Probabilistic Model}
Denoising diffusion probabilistic models (DDPMs) \cite{Ho.2020} are generative models aiming to learn the underlying data distribution $p(\bm{x})$ by reversing a forward diffusion process. 
The training of DDPMs consists of two phases: the forward phase and the reverse phase.
During the forward phase, DDPMs transform the initial data  $\bm{x}_0$ into Gaussian noise $p(\bm{x}_T) = \mathcal{N}(\mathbf{0}, \mathbf{I})$ using a predefined noising procedure. This noising procedure, also known as a noise scheduler, systematically adds Gaussian noise $\bm{\epsilon}$ at each diffusion step  $t = 1,\dots, T$ until it converges to a standard normal Gaussian noise for $T\rightarrow \infty$.
The noised data $\bm{x}_t$ at diffusion step $t$ is defined as
\begin{align}
\bm{x}_t = \sqrt{\overline{\alpha}_t} \bm{x}_{0} +   \sqrt{1-\overline{\alpha}_t}\bm{\epsilon} \quad \textrm{with } \bm{\epsilon}\sim\mathcal{N}(\bm{0}, \mathbf{I}). \label{eq:forwardDiffusion}
\end{align}
The distribution of a noised data sample $\bm{x}_t$ can be represented by $q(\bm{x}_t | \bm{x}_0) = \mathcal{N}(\sqrt{\overline{\alpha}_t} \bm{x}_{0}, (1-\overline{\alpha}_t)\mathbf{I} )$ with mean vector \mbox{$\bm{\mu}_t =\sqrt{\overline{\alpha}_t} \bm{x}_{0}$} and covariance matrix \mbox{$\bm{\Sigma}_t = (1-\overline{\alpha}_t)\mathbf{I}$}. The parameter $\overline{\alpha}_t$ results from the noise scheduler and indicates the noise level at diffusion step $t$. Although various noise scheduling strategies exist, the cosine noise scheduler introduced by Nichol \etal~\cite{Nichol.2021} has shown particularly good performance. It is defined as
\begin{align}
\overline{\alpha}_t\!=\!\frac{f(t)}{f(0)} \quad \quad \quad f(t)\!=\!\cos^2\!\left( \frac{t/T\!+\!s}{1\!+\!s}\!\cdot\!\frac{\pi}{2} \right),
\end{align}
where $s\in\mathbb{R}^+$ is a small offset, \eg $s = 0.008$, to prevent $\overline{\alpha}_t$ from being too small near $t = 0$.  The offset $s$ improves noise prediction in the early timesteps \cite{Nichol.2021}.
In the reverse phase, a neural network $p_\theta(\bm{x})$ is trained to gradually undo the transformation that occurs in the forward phase. At each step of the reverse phase, the model takes a noisy input and learns to reduce the level of noise by recovering some of the obscured information. Hence, DDPMs learn to approximate the conditional distribution $p_\theta(\bm{x}_{t-1}|\bm{x}_t, t)$ by optimizing the model parameters $\theta$. 
By repeating the statistical independent denoising step using
\begin{align}
p_\theta(\bm{x}_{0:T}) = p(\bm{x}_T) \prod_{T}^{t=1} p_\theta(\bm{x}_{t-1}|\bm{x}_t, t), \label{eq:REV}
\end{align}
the original data can effectively be recovered from the noisy data. 
\mbox{$p_\theta(\bm{x}_{t-1}|\bm{x}_t, t)=\mathcal{N}(\bm{\mu}_{\theta}(x_t, t), \bm{\Sigma}_\theta(\bm{x}_t, t))$} denotes the denoising transition step. The covariance $\bm{\Sigma}_\theta(\bm{x}_t, t)$ can either be learned or set to the variance determined by the forward diffusion $\bm{\Sigma}_\theta(\bm{x}_t, t) = \bm{\Sigma}_t$, where $\bm{\Sigma}_t=\sigma^2(t)\mathbf{I}$ and
\begin{equation}
{\sigma}^2(t)= \frac{(1-\alpha_t)(1-\overline{\alpha}_{t-1})}{1-\overline{\alpha}_t},
\end{equation}
where $\alpha_t = \frac{\overline{\alpha}_t}{\overline{\alpha}_{t-1}}$.
Instead of predicting the denoised data $\bm{\mu}_{\theta}(\bm{x}_t, t)$, the authors of \cite{Ho.2020} found that predicting the noise terms $\bm{{\epsilon}}_\theta(\bm{x}_t,t)$ is more stable. The commonly used simplified training objective results in
\begin{align}
\mathcal{L}_{\textrm{simple}} = \mathbb{E}_{t \sim [1, T]} \left[
\vert \vert
\bm{\epsilon} - \bm{\epsilon}_\theta(\bm{x}_t, t)
\vert \vert^2_2
\right].
\end{align}
The goal of DDPMs is to learn the noise that needs to be removed in each denoising step from distorted data in order to recover the original data.
Once the training has converged, new data can be generated by repeatedly computing
\begin{align}
\bm{x}_{t-1} &= \frac{1}{\sqrt{\alpha_t}} 
\left(
\bm{x}_t - \frac{1-\alpha_t}{\sqrt{1-\overline{\alpha_t}}} \bm{\epsilon}_\theta(\bm{x}_t, t)
\right) +
\bm{\Sigma}_t \bm{\epsilon},
\end{align}
where $\bm{\epsilon} \sim \mathcal{N}(\bm{0}, \mathbf{I})$.
\subsection{Classifier-Free Guidance} 
In diffusion models, the term guidance refers to controlling the generation process by incorporating additional conditions or modalities. The classifier-free guidance proposed in \cite{Ho.2022} suggests a method that does not rely on an explicit classifier to provide guidance for the diffusion model.
In this guidance approach, an unconditional noise estimator $\bm{\epsilon}_\theta(\bm{x}_t, t)$ and a conditional noise estimator $\bm{\epsilon}_\theta(\bm{x}, c, t)$ are jointly trained. Both are implemented through one neural network. Thus, the class identifier $c$ of the unconditional model is set to zero for the generation process such that $\bm{\epsilon}_\theta(\bm{x}_t, t) = \bm{\epsilon}_\theta(\bm{x}, c=0, t)$.
During sampling, the noise estimate $\bm{\tilde{\epsilon}}_\theta (\bm{x}_t, c, t)$ of the guided DDPM is determined by
\begin{equation}
\bm{\tilde{\epsilon}}_\theta (\bm{x}_t, c, t) = (1 + w)	\bm{\epsilon}_\theta (\bm{x}_t, c, t) - w	\bm{\epsilon}_\theta (\bm{x}_t,t),
\end{equation}
where $w\in \mathbb{R}$ is the guidance scale. The guidance scale is used in conditional diffusion models to balance diversity and sample fidelity. It controls how much influence the condition has over the generation process: it decreases the unconditional likelihood  with a negative score term while simultaneously increasing the conditional likelihood of a sample\cite{Ho.2022}. A higher guidance scale $w$ can lead to samples that closely match the conditioning information, resulting in higher fidelity but potentially lower diversity. Vice versa, a lower guidance scale can result in more diverse samples but with less fidelity to the conditioning information. 
\section{Method}
\label{sec:method}
\begin{figure*}[t!]
	\input{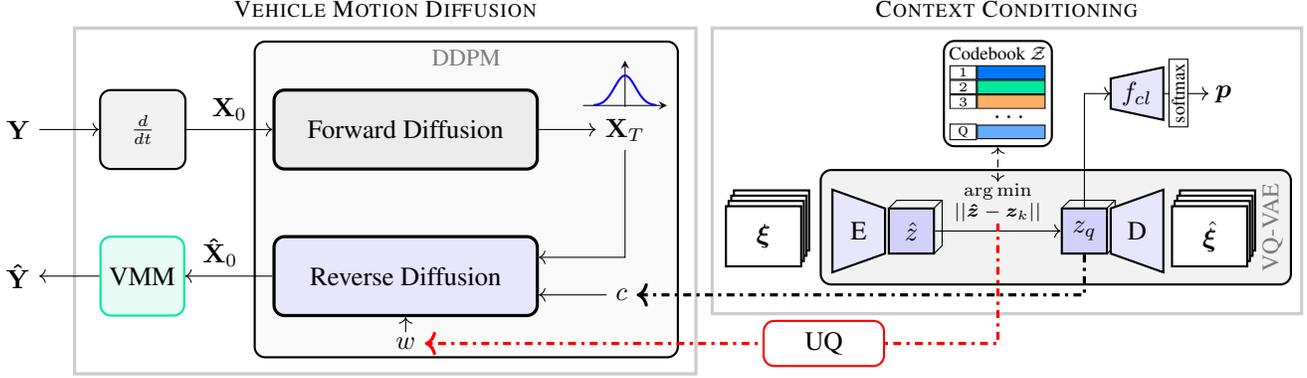}
	\vspace{-17pt}
	\caption{Architecture of cVMD, composed of three primary components: vehicle motion diffusion module, context conditioning module, and uncertainty quantification unit (UQ). The context conditioning module, realized as \mbox{VQ-VAE}, discretizes the traffic scenario $\boldsymbol{\xi}$. The index $q$ of the discretized scenario context is then passed to the diffusion model as condition $c$. UQ determines the prediction uncertainty, which is used to adaptively modify the guidance scale $w$ of the vehicle motion diffusion module used for the trajectory prediction.}
	\label{fig:vmd_architecture}
	\vspace{-10pt}
\end{figure*}

This section introduces the architecture of cVMD. As illustrated in Fig. \ref{fig:vmd_architecture}, the network consists of three main components: the vehicle motion diffusion module, the context conditioning module and the uncertainty quantification unit (UQ). The module for context conditioning captures and categorizes the scenario context, while the vehicle motion diffusion module performs the trajectory prediction. The UQ embedded in cVMD estimates the model uncertainty. This uncertainty is also used in the uncertainty-adaptive trajectory prediction. The subsequent subsections provide a detailed explanation of each component and how they are integrated within cVMD. Initially, the considered problem formulation is presented.
\subsection{Problem Formulation}
Let dataset $\mathcal{D} = \{(\boldsymbol{\xi}^{(m)}, \mathbf{Y}^{(m)}, \bm{s}^{(m)})\}^{M}_{m=1}$ be composed of $M$ distinct data samples. Each data \mbox{sample} holds the traffic scenario observations $\boldsymbol{\xi}^{(m)}\in \mathbb{R}^{N \times F \times T_\textrm{obs} }$, the future trajectory $\mathbf{Y}^{(m)} \in \mathbb{R}^{2 \times T_\textrm{pred} }$ of a selected target vehicle and the maneuver class $\mathbf{s}^{(m)} \in \mathbb{R}^{3}$. The maneuver class $\mathbf{s}^{(m)}$ is an one-hot encoded vector, indicating if the future trajectory $\mathbf{Y}^{(m)}$ is a lane change left (lcl), lane change right (lcr) or keep lane (kl) maneuver.
Based on the motion \mbox{observation $\boldsymbol{\xi}^{(m)}$} of $N=9$ vehicles within an interactive traffic scenario for the time span $T_\textrm{obs}=\SI{3}{\second}$, the task is to predict the trajectory $\mathbf{Y}^{(m)}$ of the target vehicle \mbox{$i \in \{1, \dots, N\}$}. During the observation period, a total of $F=4$ vehicle features are taken into account such that the motion information of the \mbox{$j$-th} participating vehicle is ${\boldsymbol{\xi}_j^{(m)}= [\bm{x}_{j}, \bm{y}_{j}, \bm{v}_{j,\textrm{x}}, \bm{v}_{j,\textrm{y}}]^\textrm{T}}$, containing the past longitudinal and lateral positions ($\bm{x}_{j}, \bm{y}_{j}$) and velocities ($\bm{v}_{j,\textrm{x}}, \bm{v}_{j,\textrm{y}}$) up to the current time step $ t_0 $. 
Based on the observed traffic scenario $\mathbf{\boldsymbol{\xi}}^{(m)}$, the network is tasked with predicting the trajectory of the selected targeted vehicle \mbox{${\mathbf{Y}}^{(m)} = [\bm{{x}}_\textrm{pred}, \bm{{y}}_\textrm{pred}]^\textrm{T}$}, where $\bm{{x}}_\textrm{pred}, \bm{{y}}_\textrm{pred} \in \mathbb{R}^{T_\textrm{pred}}$. The prediction horizon for the trajectory is set to $T_\textrm{pred}=\SI{5}{\second}$. 
\subsection{Context Conditioning}
The context conditioning module is used to determine and categorize the context of an observed traffic \mbox{scenario $\boldsymbol{\xi}^{(m)} \in \mathbb{R}^{N \times F \times T_\textrm{obs}}$}. The underlying goal is to discretize the space of possible scenario constellations. Although there are an infinite number of possible traffic scenario constellations, this work assumes that they can be decomposed into a discrete set of scenario representatives $q \in \{1, \dots, Q\}$. The rationale behind this is that comparable traffic scenarios lead to similar motion patterns for selected traffic participants. While similar traffic scenarios may differ in terms of the exact positioning and movement of the vehicles, they do provide a degree of context similarity. High contextual similarity between a new traffic scenario and a previously categorized scenario enhances certainty regarding the future behavior of the participants.
To put it differently, the context conditioning module performs a clustering process, whereby each \mbox{scenario $\boldsymbol{\xi}^{(m)}$} is assigned to a distinct cluster $q$ corresponding to its specific scenario context.
In this work, a VQ-VAE \cite{VQVAE.2018} is applied to interpret, categorize and cluster traffic scenarios with high contextual similarity. Given a traffic scenario observation $\boldsymbol{\xi}^{(m)}$, the VQ-VAE assigns a context category $q$ to this observation. VQ-VAE combines the concepts of Variational Autoencoders (VAEs) \cite{vae2022} and Vector Quantization (VQ). The architecture consists of an encoder $E$ that maps input data $\boldsymbol{\xi}^{(m)}$ to a latent representation \mbox{$\bm{\hat{z}}^{(m)}\in \mathbb{R}^{R_q}$} with the vector dimension $R_q$ and a decoder $D$ that reconstructs the input data from the latent representation. In VQ-VAE, the latent space is discretized, $\bm{z}_q^{(m)} = f_q(\bm{\hat{z}}^{(m)})$, prior to being fed into the decoder, rather than directly reconstructing the original input data. During quantization, the latent representation is mapped to a single codebook vector $\bm{\hat{z}}^{(m)} \rightarrow \bm{z}_q^{(m)}$ from a finite set of codebook vectors $\mathcal{Z} = \{\bm{z}_1, . . . , \bm{z}_{Q}\}$, using the Euclidean distance
\begin{equation}
\bm{z}_q^{(m)}  = f_q(\bm{\hat{z}}^{(m)}) = \arg \min_{z_k \in \mathcal{Z}} \vert \vert \hat{\bm{z}}^{(m)} - \bm{z}_k \vert \vert^2_2 \label{eq:vaequanitze},
\end{equation}
with $\bm{z}_k \in \mathbb{R}^{R_q}$.
Subsequently, the decoder tries to reconstruct the input data based on the determined codebook entry $\hat{\bm{\xi}}^{(m)} = D(\bm{z}_q^{(m)})$. During training, the model parameters and codebook vectors $\mathcal{Z}$ are optimized via 
\begin{equation}
\begin{aligned}
\mathcal{L}_\textrm{vq} = ||\boldsymbol{\xi}^{(m)} - \hat{\boldsymbol{\xi}}^{(m)} ||^2 &+ \vert \vert \textrm{sg}[E(\boldsymbol{\xi}^{(m)})] - \bm{z}_q^{(m)} \vert \vert^2_2\\ &+ \vert \vert \textrm{sg}[\bm{z}^{(m)}_q] - E(\boldsymbol{\xi}^{(m)}) \vert \vert^2_2,
\end{aligned}
\label{eq:vqganloss}
\end{equation}
where $\textrm{sg}[\cdot]$ denotes the stop-gradient operation.
In this work, the loss function of the VQ-VAE (Eq. \ref{eq:vqganloss}) is extended to include a classification task in the latent space. To highlight the details in the constellations of a traffic scenario that lead to different following maneuvers (lcl, lcr, kl), a linear classifier $f_{cl}(\bm{z}_q^{(m)})$ is added. The classifier assigns a maneuver class to each selected codebook entry $\bm{z}_q^{(m)}$. To penalize false classification a cross-entropy loss
\vspace{-2pt}
\begin{align}
\mathcal{L}_\textrm{cl} &= - \sum^{S}_{i=1} {s}_i^{(m)} \log({p}_i^{(m)}),\\
\bm{p}^{(m)} &= \textrm{softmax}(f_{cl}(\bm{z}_q)),
\vspace{-2pt}
\end{align}
\vspace{-1pt}
is applied, where $\bm{s}^{(m)}\in\mathbb{R}^S$ is the ground truth label for the $S=3$ different maneuver classes (lcl, kl, lcr) and the predicted class $\bm{p}^{(m)} \in \mathbb{R}^S$. With the introduction of the loss $\mathcal{L}_\textrm{cl}$, the latent space is additionally forced to form meaningful codebook entries.
The complete objective for training the context conditioning architecture reads
\begin{equation}
\mathcal{L}_\textrm{cc} = \mathcal{L}_\textrm{vq} + \lambda \mathcal{L}_\textrm{cl},
\vspace{-2pt}
\end{equation}
where $\lambda$ is an adaptive weight.

Once the training of the VQ-VAE has converged, each codebook entry $\bm{z}_q \in \mathcal{Z}$ is an embedding for a specific traffic scenario context. Thereby, the vast space of scenario constellations is divided into $Q$ distinct scenario clusters, with each cluster $q$ representing a similar scenario context.

\subsection{Vehicle Motion Diffusion (VMD)}
The aim of the VMD module is to model and predict feasible patterns of vehicle motion. The trajectory prediction task is performed by classifier-free guided DDPM. Unlike the context conditioning module, this module does not have access to the observed scenario context $\bm{\xi}^{(m)}$. Instead, it only receives the scenario context index \mbox{$c^{(m)} = q$} as an input condition for the DDPM. When presented with condition $c^{(m)}$, the VMD module generates a context-adaptive future trajectory prediction \mbox{${\mathbf{\hat{Y}}}^{(m)} = [\bm{{\hat{x}}}_\textrm{pred}, \bm{{\hat{y}}}_\textrm{pred}]^\textrm{T}$} for the selected target vehicle. 
However, the DDPM does not directly predict the trajectory coordinates $(\bm{{\hat{x}}}_\textrm{pred}, \bm{{\hat{y}}}_\textrm{pred})$ as is common in most approaches. Instead, it learns to predict a sequence of motion parameters $\mathbf{\hat{X}}_0^{(m)}$ for a Vehicle Motion Model (VMM). The VMM transforms the motion parameters $\mathbf{\hat{X}}_0^{(m)}$ into trajectory ${\mathbf{\hat{Y}}}^{(m)}$.
\subsubsection{DDPM}
A classifier-free guided DDPM is utilized to forecast the sequence of motion parameters $\mathbf{\hat{X}}_0^{(m)}$. The DDPM is implemented as described in the preliminaries. The forward diffusion process has no learnable parameters, whereas the reverse diffusion process is approximated utilizing \mbox{U-Net \cite{Ronneberger.15}} architecture.
During the training phase, DDPM learns which trajectories can be followed, or are likely to be followed, for a scenario context \mbox{$c^{(m)}\!=\!q$}. The condition $c^{(m)}$ is a guidance modality to generate scenario-dependent trajectory predictions. Due to the discretization of the scenario context, however, DDPM is not trained with a single most likely trajectory for each scenario $q$, but rather with a set of possible trajectories. Depending on the total number of scenario representatives $Q$, the possible trajectories for $q$ can vary greatly (for a low value of $Q$) or hardly at all (for a high value of $Q$). The rationale behind this is, that comparable traffic scenarios lead to similar motion patterns for a given target vehicle. However, the exact execution of the motion pattern can vary. Even in identical scenarios, different drivers will respond with distinct maneuvers or trajectories. The DDPM is able to represent the inherent uncertainty in predicting trajectories, which is challenging to achieve using discriminative ML architectures. Although this approach may result in lower performance in traditional metrics, such as average error between the predicted the trajectory and ground truth, it has the advantage of capturing the inherent stochasticity. \\  
During the application phase, a trajectory prediction is generated based on the given condition $c$ using only the reverse diffusion path. The forward diffusion path is not necessary and therefore discarded. In this phase, however, the network additionally considers the guidance scale $w$. This hyperparameter determines how much influence the context condition has on the trajectory prediction. In general, the guidance scale $w$ is a fixed value. In this work, however, the parameter is introduced as variable and is parameterized based on an estimate of the model's uncertainty of the trajectory prediction. The more confident the model is that it has seen a similar scenario before, the higher its prediction confidence and the higher its guide scale $w$. A detailed explanation of the realization is explained in Sec. \ref{subsec:uc}.
\subsubsection{Vehicle Motion Model (VMM)}
\begin{figure}[t!]
	\centering
	\begin{tikzpicture}
	\def\mu{0};
	\def\var{1.0};
	\def\h{0.07*gauss(\mu,\mu,\var)};
	\def\qzero{\mu-3*\var};
	\def\qone{\mu-1.8*\var};
	\def\qtwo{\mu+1.8*\var};
	\def\qthree{\mu+3*\var};
	\begin{axis}[
	anchor={center},
	axis lines=center,
	axis line style=thick,
	xlabel={$x_0$},
	ylabel={$p(x_0)$ },
	enlargelimits=false, 
	ticks=none,
	width=5cm,
	height=3cm,
	xmin=-10, xmax=10,
	ymin=-0.01, ymax=0.45,
	clip=false, 
	name= orginalXY,
	]
	\addplot[blue, densely dashed, thick,name path=normaldist,domain={-8}:{8},samples=100,smooth,] {0.25*(gauss(x,-3.5,1.2*\var) + 
		gauss(x,\mu,0.35*\var) +
		gauss(x,1.5,1.25*\var)+
		gauss(x,5,0.75*\var))}; \label{distPos}
	\addplot[red,thick,name path=normaldist,domain={-8}:{8},samples=100,smooth,] {0.25*(gauss(x,\mu,1.2*\var) + 
		gauss(x,\mu+1.3,0.9*\var) +
		gauss(x,\mu-1.5,1.55*\var)+
		gauss(x,\mu-0.2,1.75*\var))}; \label{distOP}
	\addplot [domain=-3.5:3.5, samples=100, name path=xaxis, thin, color=white, opacity=0] {0*x +0.0025};	
	\end{axis}
	\begin{axis}[
	at={(4.3cm, 0)},
	anchor={center},
	axis lines=center,
	axis line style=thick,
	xlabel={$x_T$},
	ylabel={$p(x_T)$ },
	enlargelimits=false, 
	ticks=none,
	width=5cm,
	height=3cm,
	xmin=-8, xmax=8,
	ymin=-0.01, ymax=0.6,
	clip=false,
	name= gaussDIFF,
	]
	\addplot[blue,thick,name path=normaldist,domain={-4}:{4},samples=100,smooth,] {gauss(x,\mu,\var))}; \label{stdplot}
	\addplot [domain=-3.5:3.5, samples=100, name path=xaxis, thin, color=white, opacity=0] {0*x +0.0025};	
	\end{axis}
	
	\begin{scope}[every node/.style={text width=2cm,align=center}]
	\node [below] at (orginalXY.south) {{(a)}};
	\node [below] at (gaussDIFF.south) {{(b)}};
	\end{scope}
	\draw [->,thin,color=black] (orginalXY.east) -- (gaussDIFF.west) node [pos=.5, above, align =center, text width = 1cm, font=\scriptsize ] (diff) {diffusion};
	\end{tikzpicture}
	\vspace{-9pt}
	\caption[Caption in ToC]{
		Diffusion processes transform (a) data $p(x_0)$ into to Gaussian noise (b) $p(x_T)$ (\ref{stdplot}). While the distribution of positional data (\ref{distPos}) is highly depending on the map data, the motion parameter distribution (\ref{distOP}) is more likely to resemble a Gaussian distribution.}
	\label{fig:datadist}
	\vspace{-12pt}
\end{figure}
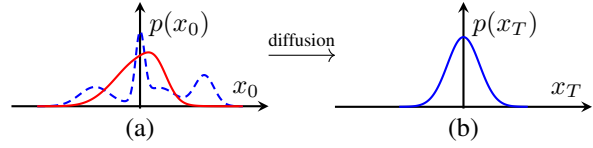

VMMs are mathematical representations of the motion kinematics of a vehicle. These models aim to capture the relationship between the vehicle's inputs and its resulting motion by considering underlying non-holonomic constraints. In this work, a VMM with variable yaw rate $\dot{\psi}_t$ and longitudinal acceleration $a_{x,t}$ is used to represent the kinematics of vehicles. As described in \cite{Botsch.2020}, the position $(x_{t}, y_{t})$, velocity $v_{t}$ and heading $\psi_{t}$ of a vehicle at time step $t+\tau$ using this specific VMM are determined computing
{
\thinmuskip=1mu
\medmuskip=2mu plus 1mu minus 2mu
\thickmuskip=1mu plus 1mu
\begin{align}
x_{t+\tau}\!&=\!x_{t}\!+\!v_{t}c(\psi_t)\tau\!+\!( a_{x,t} c(\psi_t)\!-\!\dot{\psi}_t v_t s(\psi_t))\frac{\tau^2}{2} \label{eq:VMM1}\\	
y_{t+\tau}\!&=\!y_{t}\!+\!v_{t}s(\psi_t)v\!+\!(a_{x,t}s(\psi_t)\!+\!\dot{\psi}_t v_t c(\psi_t))\frac{\tau^2}{2}\\
v_{t+\tau}\!&=\!v_{t}\!+\!a_{x,t}\tau\\
\psi_{t+\tau}\!&=\!\psi_{t}\!+\!\dot{\psi}_t \tau \label{eq:VMM2}
\end{align}}
where $c(\psi_t)\!=\!\cos(\psi_t)$, $s(\psi_t)\!=\!\sin(\psi_t)$ and time increment $\tau$. For the used VMM the motion parameters are defined as \mbox{$\mathbf{{X}}_0^{(m)}= [{\dot{\bm{\psi}}}, {\bm{a}}_{x}]$}, with \mbox{${\dot{\bm{\psi}}},{\bm{a}}_{x} \in \mathbb{R}^{T_\textrm{pred}}$}.
The ground truth motion parameters \mbox{$\mathbf{{X}}_0^{(m)}= [{\dot{\bm{\psi}}, \bm{a}_{x}}]$} are calculated by the numerical derivations \mbox{$\bm{a}_x = \frac{d^2 \bm{x}_\textrm{pred}}{d t^2}$} and \mbox{$\dot{\bm{\psi}} = \arctan \left(\frac{d \bm{y}_\textrm{pred}}{dt}, \frac{d \bm{x}_\textrm{pred}}{dt}\right)$}. Thus, during training, the DDPM learns to predict the sequence of motion parameters \mbox{$\mathbf{\hat{X}}_0^{(m)} =[\hat{\dot{\bm{\psi}}},\hat{\bm{a}}_{x}]$}. Note that these motion parameters have known physical limits (${\dot{\bm{\psi}}}_\textrm{max} = \pm \SI[quotient-mode=fraction]{71.26}{\deg/\second}$\cite{Kontos.23}, ${\bm{a}}_{x, \textrm{max}} = \pm \SI[quotient-mode=fraction]{9}{\meter/\second^2}$\cite{Yusof.2016}), that are taken into account during trajectory prediction. Bounding each prediction to these physical limits ensures that the values do not exceed defined limits and that the predicted trajectory can be executed by a vehicle. The equations \mbox{Eq. (\ref{eq:VMM1})-(\ref{eq:VMM2})} incorporate the non-holonomic constraints of vehicles, relate the vehicle's motion parameters to its motion, and allow reliable prediction of vehicle trajectories.
The use of motion parameters as predicted quantities introduces an additional benefit to the learning process of the DDPM. 
As explained initially, the idea of diffusion models is to transform the data distribution gradually into a Gaussian distribution, and then learn how to reverse this process. While the distributions of acceleration and yaw rate values are likely to be very approximate to a Gaussian distribution, this is less likely to be the case for trajectory position data. This is due to the fact that the distribution of trajectory position data is highly dependent on the dataset used and the conditions of the infrastructure. If the original data distribution is already Gaussian, it simplifies both the forward and backward process. Hence, also the learning task is simpler since finding a reverse mapping from complex data distributions back to a Gaussian distribution is often challenging. The intuition behind this is illustrated in Fig. \ref{fig:datadist}.
\subsection{Uncertainty Quantification}
\label{subsec:uc}
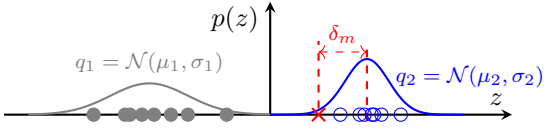
\begin{figure}[t!]
	\centering
	\begin{tikzpicture}
	\def\var{1.0};
	\begin{axis}[
	anchor={center},
	axis lines=center,
	axis line style=thick,
	xlabel={$z$},
	ylabel={$p(z)$ },
	y label style={at={(axis description cs:0.52,.85)}, anchor=east},
	enlargelimits=false, 
	ticks=none,
	width=\columnwidth,
	height=3.1cm,
	xmin=-11, xmax=10,
	ymin=-0.01, ymax=0.9,
	clip=false
	]
	\addplot[blue, thick,name path=normaldist,domain={0}:{8},samples=100,smooth,] {gauss(x, 4, 0.9)} node[right,pos=0.6] {\footnotesize{$q_2=\mathcal{N}(\mu_2, \sigma_2)$}};
	\addplot[gray, thick,name path=normaldist,domain={-10}:{0},samples=100,smooth,] {gauss(x,-5, 1.6)} node[above,pos=0.5] {\footnotesize{$q_1=\mathcal{N}(\mu_1, \sigma_1)$}};
	\addplot [domain=-3.5:3.5, samples=100, name path=xaxis, thin, color=white, opacity=0] {0*x +0.0025};
	
	\addplot [gray, only marks, mark size=2.5pt] table {
		-3.4  0
		-4.1  0
		-7.3  0
		-5.3  0
		-6.0  0
		-1.8  0  
		-5.8  0  
		-4.8  0
	}; \label{hq1}
	\addplot [blue, only marks, mark=o, mark size=2.5pt] table {
		4.20  0   
		3.90  0 
		4.60  0 
		4.30  0 
		3.70  0
		2.90  0
		5.40  0 
	}; \label{hq2}
	\addplot [red, only marks, mark=x, thick, mark size=3.5pt] table {
	2  0   
	}; \label{new_sample}
	\addplot[mark=none, thick, red, dashed] coordinates {(4, -0.05) (4,0.58)};
	\addplot[mark=none, thick, red, dashed] coordinates {(2, 0.05) (2,0.588)};
	\addplot[<->, mark=none, red, dashed] coordinates {(2, 0.5) (4, 0.5)} node[above, pos=0.5] {\footnotesize $\delta_m$};
	\end{axis}
	\end{tikzpicture}
	\vspace{-4pt}
	\caption[Caption in ToC]{Example of univariate MLE based on a set of samples $\mathcal{H}_1$(\ref{hq1}) and $\mathcal{H}_2$(\ref{hq2}). A new data sample (\ref{new_sample}) is assigned to  $q_2$ with the quantified uncertainty $\delta_m$.}
	\label{fig:MLE}
	\vspace{-12pt}
\end{figure}

Quantifying the uncertainty in a model's predictions and determining the level of confidence in those predictions is extremely important in safety-critical applications. Operating with a false sense of certainty can be hazardous and wrong predictions can have severe consequences.
Therefore, the identification of model limitations and the assessment of model uncertainty is a significant aspect of this work.
Previous studies \cite{Bohm.19, Joppich.21} have already shown that it is possible to obtain a measure of model uncertainty in the latent representation within encoder-decoder architectures. Based on these findings, in this work the prediction uncertainty of the model $\delta_m$ is estimated using Maximum Likelihood Estimation (MLE) in the latent space of the VQ-VAE and inferential statistics. The resulting model uncertainty is incorporated into the prediction of the vehicle's trajectory.
As previously explained, the idea of VQ-VAEs is to map similar scenarios in close proximity within the latent space. The associated codebook entry $\bm{z}_q^{(m)}$ for each sample embedding $\hat{\bm{z}}^{(m)}$ can be interpreted as representative of the respective scenario context. Thus, the distance of a new data point from the codebook entry in the latent space indicates the similarity to the respective scenario context representative. Once the training procedure of the VQ-VAE has converged, each data sample within $\mathcal{D}_{\textrm{train}}$ is ultimately assigned to the closest codebook entry ${\bm{z}}_q^{(m)}$ according to Eq. \ref{eq:vaequanitze}. This generates a set of assigned samples \mbox{$\mathcal{H}_q = \{ {\hat{\bm{z}}}^{(1)}, \hat{\bm{z}}^{(2)},\dots, \hat{\bm{z}}^{(h_q)} \}$} for each codebook entry \mbox{$q= 1, \dots, Q$}, where $h_q$ is the total number of samples assigned to $q$. Each set $\mathcal{H}_q $ is used to approximate the true class conditional distribution in the latent space using a tractable distribution from within a variational family $\mathcal{Q}$. In this work, all class conditional distributions $q_q \in \mathcal{Q}$ are assumed to follow multivariate Gaussian distributions ${q}_{q}(\bm{z})=\mathcal{N}(\bm{\mu}_q,\bm{\Sigma}_q)$ with mean $\bm{\mu}_q$ and covariance $\bm{\Sigma}_q$. The distribution's parameters are defined as
\begin{align}
\vspace{-3pt}
\bm{\mu}_q &=  \bm{z}_q\\
\bm{\Sigma}_q &= \mathbb{E}[(\hat{\bm{z}}^{(h)}-\bm{\mu}_q)(\hat{\bm{z}}^{(h)}-\bm{\mu}_q)^\textrm{T} ],
\end{align}
where $\hat{\bm{z}}^{(h)}\in\mathcal{H}_q$.
After using MLE to fit the class probability distributions as exemplified in Fig. \ref{fig:MLE}, the likelihood of a new observation \mbox{$\bm{\hat{z}}^{(m)} \in \mathcal{D}_\text{test}$} under the fitted model can be identified and the model uncertainty $\delta_m$ can be estimated.
Similar to \cite{Joppich.21}, the model uncertainty $\delta_m$ is quantified using the Mahalanobis distance (M-distance) 
\begin{align}
\delta_m(q_q, \hat{\bm{z}}^{(m)})\!=\!\sqrt{\left(\bm{\mu}_q\!-\!\hat{\bm{z}}^{(m)}\right)^\textrm{T} \bm{\Sigma_q}^{-1}  \left(\bm{\mu}_q\!-\!\hat{\bm{z}}^{(m)}\right)}.
\end{align}
Given a sample $\hat{\bm{z}}^{(m)}$, the M-distance evaluates the distance of the samples to the class conditional distribution ${q}_{q}(\bm{z})$. If the model uncertainty is above a certain threshold $t_c$, the observation could be classified as an outlier. This means that the current scenario is unlikely to be appropriately represented by any codebook entry and is therefore a potentially unknown scenario. Vice versa, the lower the \mbox{M-distance}, the more confident the model is in its prediction.
In the context of the introduced cVMD, model uncertainty plays an important role in the prediction of the vehicle trajectory. On the one hand, a high model uncertainty indicates that the model limits have been exceeded and a reliable trajectory prediction cannot be guaranteed. On the other hand, the uncertainty quantification can be used to adaptively parameterize the guidance scale $w$ of the diffusion model to influence the trajectory generation process.
\subsection{Uncertainty-adaptive Guidance Scale}\label{sec:UC}
The guidance scale $w$ controls to what extend the trajectory prediction process is conditioned on the provided context condition of the scenario. The higher the value, the more the model amplifies the provided condition to predict the trajectory. However, this does not mean that the value should always be set to maximum, as more guidance means less diversity and quality. A high level of guidance reduces the variety in the trajectory predictions and may create a risk of overemphasizing the condition in the generation process.
In this work, the guidance scale is therefore calculated adaptively, based on the model's identified prediction uncertainty $\delta_m$, using
\begin{equation}
w = w_\textrm{min} + \left(1- \frac{\min(\delta_m, t_c)}{t_c}\right) \left(w_\textrm{max} - w_\textrm{min}\right), \label{eq:adapw}
\end{equation}
where $w_\textrm{min}, w_\textrm{max} \in \mathbb{R}$ are the minimal and maximal parameters for $w$.
According to Eq. \ref{eq:adapw}, when the model uncertainty is low, the guidance scale is consequently large. As a result, the model's generation of the trajectory prediction is strongly conditioned on the scenario context. This setting implies that there is a comprehensive understanding of the existing context condition, as similar scenarios have been encountered before. 
However, due to the non-deterministic nature of diffusion models, each trajectory generation process inherently embodies a degree of stochasticity.
This reflects the real-world principle that the same maneuver can be performed in many different ways due to individual driving behaviors.
Conversely, if there is a high model uncertainty, the future trajectory is less certain, suggesting that the model has not been previously exposed to a similar situation. The resulting lower guidance scale leads to a trajectory prediction with more variability, implying reduced prediction reliability. 

\section{Dataset and Experiments}
\label{sec:experiments}
In this work, the performance of the proposed cVMD architecture is experimentally evaluated and compared with state-of-the-art models. In addition, an ablation study for the parameterization of guidance scale $w$ is performed to evaluate the optimal hyperparameter configuration. Since the performance of the overall architecture is sensitive to the discretization quality of the VQ-VAE, also the robustness of the context conditioning is investigated. \\
\textbf{Dataset. \,}
For the experiments, the publicly available highway dataset highD~\cite{highD} is used due to its extent of application-oriented scenarios. The highD dataset contains drone recordings of German highways taken at a frequency of \SI{25}{\hertz}. The dataset naturally contains a large imbalance of scenarios, as lane changes occur less frequently than lane keeping. Therefore, it is pre-processed in such a way that the extracted scenarios are distributed in a uniform manner. The resulting data format is consistent with the previously explained problem definition. The extracted scenarios are split into the subsets $\mathcal{D}_{\textrm{train}}$ (9,841 samples for training) and $\mathcal{D}_{\textrm{test}}$ (4,217 samples for testing and experiments).\\
\textbf{Implementation Details. \,}
\renewcommand*{\thefootnote}{\fnsymbol{footnote}}
\renewcommand\footnoterule{\rule{0.4\linewidth}{0.8pt}}
The training processes of the vehicle motion diffusion module and the context conditioning module are decoupled. First, the context conditioning module is trained with batch size \mbox{$B_{1}=64$}, learning rate \mbox{$lr_1=$\SI{4.5e-6}{}}  and \mbox{$\lambda=1$} for a total number of epochs \mbox{$E_{1}=1200$}. The VQ-VAE codebook is configured with \mbox{$Q = 60$} entries, where each codebook entry is of dimension \mbox{$\bm{z}_q \in \mathbb{R}^{64}$}. Once the training procedure for the VQ-VAE is completed, its parameters are fixed. Secondly, the vehicle motion diffusion module is trained with batch size \mbox{$B_{2}=64$} and learning rate \mbox{$lr_{2}=$\SI{1.0e-4}{}} for \mbox{$E_{2}=50$} epochs. For details of architecture implementation and code see: \url{https://github.com/mb-team-thi/conditioned-vehicle-motion-diffusion}.
\section{Evaluation}
\label{sec:evaluation}
\subsection{Codebook entry}
\begin{figure}[t]
	\centering
	\vspace{1pt}
	\begin{tikzpicture}
	\begin{axis}[
	ybar stacked,
	ymajorgrids,
	grid style={dashed,gray!30},
	bar width = 3pt,
	height=3.0cm,
	xtick align=center,
	width=0.95*\linewidth,
	xlabel={\small $q$},
	ylabel={\small $\log_2(\#)$},
	x label style={at={(axis description cs:0.5,+0.045)}, anchor=north},
	y label style={at={(axis description cs:0.12,.5)}, anchor=south},
	xtick={0,10,20,30,40,50,60},
	xticklabels={0,10,20,30,40,50,60},
	table/col sep=comma,
	ymin=0,
	xmin=-0.5,
	xmax=60,
	legend entries={ \small lcl, \small kl,\small lcr},
	legend style={at={(1,0.5)}, xshift = 0.55cm, 
		anchor=center, nodes=right,  minimum size=0.2cm, inner sep=1pt},
	legend columns=1,
	]
	\addplot+ table[x =idx, y=lcl] {data/e1062_train_loghist.csv};
	\addplot+ table[x =idx, y=kl] {data/e1062_train_loghist.csv};
	\addplot+ table[x =idx, y=lcr] {data/e1062_train_loghist.csv};
	\end{axis}
	\end{tikzpicture}
	\vspace{-21pt}
	\caption{Stacked histogram for the selected context condition $q$ from the codebook. The histogram has a logarithmic scale.}
	\label{fig:stackedhistogramTrain}
	\vspace{-8pt}
\end{figure}
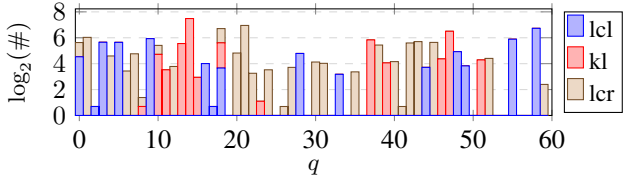
The VQ-VAE discretizes the infinite scenario space by assigning each scenario embedding $\hat{\bm{z}}^{(m)}$ a specific context condition, denoted as $q$. A visual representation of the distribution of the context indices that are assigned to the samples of $\mathcal{D}_\textrm{train}$ on the basis of Eq. \ref{eq:vaequanitze} is given in Fig. \ref{fig:stackedhistogramTrain}. The color of a bar indicates the maneuver class (lcl, kl, lcr) of the target agent's future trajectory according to the scenario context. If a single bar is composed of different color segments, it means that one categorized scenario context leads to different maneuver classes. Ideally, however, each bar should be represented by a single color. This way, there is no ambiguity as to which maneuver category is going to be performed by the target vehicle. As can be seen, the VQ-VAE training process resulted in the maneuver converging using $Q_t=49$ of the $Q=60$ entries. 
From the stacked histogram, it is noticeable that most traffic scenario contexts $q$ are followed by a typical target agent maneuver. Yet, some scenario types, \eg $q\!=\!18$, have no clear following maneuver. This means that in certain scenario constellations, the target vehicle reacted with different maneuvers. To quantify the level of maneuver diversity for the context conditions, average Shannon entropy $H_\textrm{avg} = \mathbb{E}_{q=1,\dots,Q}[H_\textrm{q}]$ is calculated. The entropy $H_\textrm{q}$ for condition $q$ is
\vspace{-3.3pt}
\begin{equation}
H_\textrm{q} = - \sum_{i=1}^{S} p_i \log_2 p_i,
\vspace{-3.3pt}
\end{equation} 
where $p_i$ is the probability of the maneuver class $i$ being assigned to condition $q$ and $S=3$. Thus, $H_\textrm{q}$ is a measure of the unpredictability of the maneuver class for the context condition $q$. As there are three potential classes $S$, Shannon entropy can range from $0$ (complete purity) to $\log_2(3)=1.585$ (complete impurity, instances are evenly distributed among all classes).
Computing the Shannon entropy separately for the training and test datasets resulted in a significantly lower entropy for the training dataset $H_\textrm{avg}(\mathcal{D}_\textrm{train}) = 0.01$ in comparison to the test dataset $H_\textrm{avg}(\mathcal{D}_\textrm{test}) = 0.39$.
Hence, on average, the distribution of maneuver classes for a context condition $q$ is more impure for $\mathcal{D}_\textrm{test}$ than $\mathcal{D}_\textrm{train}$. While this is an indication that the VQ-VAE did not learn to generalize effectively, the entropy value of $0.39$ still indicates that there is a relative majority of one class for each context condition $q$. However, since the model's ability to correctly predict future trajectories is only as robust as the capacity of the clustering algorithm, future work is aimed at improving clustering performance in terms of clear scenario differentiation and generating more appropriate codebook entries. Nevertheless, the current latent space of the embeddings $\hat{\bm{z}}^{(m)}$ and the generated clusters can be thoroughly examined using the visualization tool \texttt{VQSPEC}\footnote{\url{https://mb-team-thi.github.io/VQSPEC/}}, based on \cite{umapexplorer2019, pca1993, TSNE, umap_dimred}.
\begin{table}[t!]
	\centering
	\vspace{7pt}
	\begin{tabular}{c|c|c|c|c|}
		\multicolumn{1}{c|}{Ablation}&
		\multicolumn{2}{c|}{highD} \\
		$w$ &ADE [\SI{}{\meter}] & FDE [\SI{}{\meter}]@\SI{5}{\second} \\
		\hline
		\hline
		1  	    & 1.90 & 4.02 \\	\hline 
		3  		& 1.85 & 3.90 \\	\hline 
		5  		& 1.82 & 3.82 \\	\hline 
		7  		& 1.88 & 3.92 \\	\hline 
		13  	& 1.93 & 4.01 \\	\hline 
		$\mathit{\bm{uc}}$		& \textbf{1.79} & \textbf{3.76} \\	\hline 
	\end{tabular}%
\vspace{-5pt}
	\caption{Ablation study results showing influence of guidance scale $w$ on the trajectory prediction performance.}
	\label{tab:ablations}
\vspace{-10pt}
\end{table}
\subsection{Ablation study}
The ablation study evaluates the importance of the hyperparameter $w$ within the predictive diffusion model. Tab. \ref{tab:ablations} shows the results of the trajectory prediction performance as the value of parameter $w$ is varied. Similar to \cite{Neumeier.2022}, the Average Displacement Error (ADE) and the Final Displacement Error (FDE) at $T_\textrm{pred}=\SI{5}{\second}$ are used to evaluate performance in the vehicle trajectory prediction task. In contrast to the settings $w =\{1,3,5,7,10,13\}$,  setting $w=uc$ indicates the uncertainty-adaptive computation of guidance scale $w$ according to Eq. \ref{eq:adapw}. The hyperparameters are set to $t_c =10$, $w_\textrm{min} = 1$ and $w_\textrm{max} = 7$.
In the conducted ablation study, the prediction performance improved progressively when the guidance scale was increased from $w=1$ to $w=5$. Increasing the guiding scale beyond $w=5$ leads to a gradual decrease in performance. Thus, based on the evaluation metrics used, $w=5$ results in the best performing prediction model when using a fixed guidance scale.
However, the overall best performing prediction was achieved when setting the guidance scale uncertainty-adaptive. This highlights the importance and effectiveness of the proposed approach in setting the guidance scale as a function of model uncertainty, thereby managing the fidelity-diversity trade-off in the diffusion model generation process. Information on uncertainty can help assess and influence the level of confidence a model has in its trajectory predictions. 
\begin{table}[t!]
	\centering
	\vspace{4pt}
	\begin{tabular}{c|c|c|}
		\multirow{2}{*}{Architecture}&
		\multicolumn{2}{c|}{highD} \\
		&ADE [\SI{}{\meter}] 
		& FDE [\SI{}{\meter}]@\SI{5}{\second}  \\
		\hline
		\hline
		GFTNNv2 \cite{Neumeier.ITSC2023}&  \textbf{0.72} & \textbf{1.80}  \\	\hline 
		HSTA \cite{HSTA.2021}        	& 2.18 & 4.56 	\\	\hline
		CS-LSTM	 \cite{Deo.2018}	 	& 2.88 & 5.71\\	\hline 
		MHA-LSTM(+f) \cite{Messaoud.2021} & 2.58 & 5.44 	\\	\hline 
		Two-channel \cite{Mo.08.07.2021}  & 2.97 & 6.30 	\\	\hline 
		RA-GAT \cite{Ding.2021} & 3.46 & 6.93 \\ \hline 
		\textbf{cVMD ($\bm{\mathit{w=uc}}$)}   & 1.79 & 3.76 \\ \hline 
	\end{tabular}%
\vspace{-5pt}
	\caption{Prediction performance of different state-of-the-art architectures based on the metrics ADE and FDE.}
	\label{tab:comparison}
	\vspace{-7pt}
\end{table}

\begin{figure}[t]
\centering
\begin{tikzpicture}
\def \hvlen {6.80};
\def \hvlenmin {13.80};
\def \hvlenmax {2.80};
\def \hvwidth {1.051};
\def \lanewidth {3.4/2};

\begin{axis}[
table/col sep=comma,
xlabel={$x [m]$},
ylabel={$y [m]$ },
width=\columnwidth,
height=3.0cm,
xtick={0,20,40,60,80,100, 120, 140, 160},
xticklabels={0,20,40,60,80,100,120, 140, 160},
table/col sep=comma,
ymax = 3,
ymin =-5,
xmin = -\hvlenmin-2,
xmax = 151,
x label style={at={(axis description cs:0.5,+0.1)}, anchor=north},
y label style={at={(axis description cs:0.1,.5)}, anchor=south},
]
\addplot[gray, samples=100, domain=-25:160, dashed, line width=0.75mm, dash pattern={on 18pt off 28pt}] {\lanewidth}; 
\addplot[gray, samples=100, domain=-25:160, dashed, line width=0.75mm, dash pattern={on 18pt off 28pt}] {-\lanewidth}; 

\addplot [name path=upper3, fill=none, draw=none] table [x=X, y expr= \thisrow{uplim3}] {data/prediction_interval_idx4.csv};
\addplot [name path=lower3, fill=none, draw=none] table [x=X, y expr= \thisrow{lowlim3}] {data/prediction_interval_idx4.csv};

\addplot [name path=upper2, fill=none, draw=none] table [x=X, y expr= \thisrow{uplim2}] {data/prediction_interval_idx4.csv}; 
\addplot [name path=lower2, fill=none, draw=none] table [x=X, y expr= \thisrow{lowlim2}] {data/prediction_interval_idx4.csv};

\addplot [name path=upper, fill=none, draw=none] table [x=X, y expr= \thisrow{uplim}] {data/prediction_interval_idx4.csv}; 
\addplot [name path=lower, fill=none, draw=none] table [x=X, y expr=\thisrow{lowlim}] {data/prediction_interval_idx4.csv};
\addplot[blue!20] fill between[of=lower3 and upper3]; \label{sigma3}
\addplot[blue!50] fill between[of=lower2 and upper2];\label{sigma2}
\addplot[blue!90] fill between[of=lower and upper];\label{sigma1}

\addplot [cyan, thick] table [x=X, y=Y]{data/prediction_interval_idx4.csv};  \label{mean}
\addplot [very thick, draw=red, dotted] table [x=x0, y=y0]{data/idx4gt.csv};  \label{gt}
\addplot[] graphics[xmin=-\hvlenmin, ymin=-\hvwidth, xmax=\hvlenmax, ymax=\hvwidth]{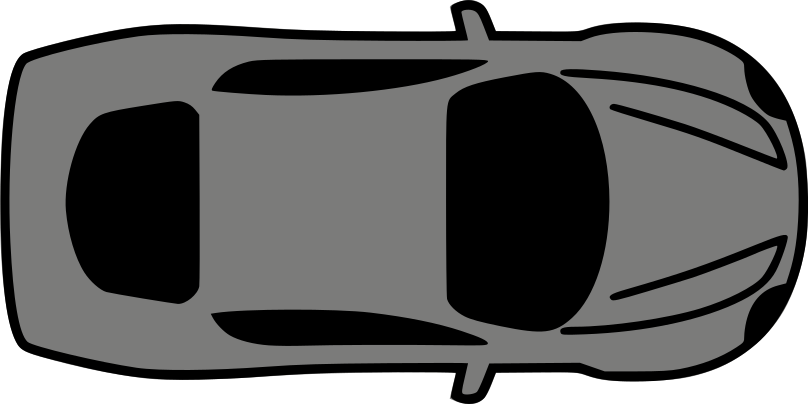};

\end{axis}
\end{tikzpicture}
\vspace{-11pt}
\caption[Caption in ToC]{Trajectory prediction $\bm{\mu}_p$ (\ref{mean}) with confidence intervals $\bm{\sigma}_p$ (\ref{sigma1}), $2\bm{\sigma}_p$ (\ref{sigma2}), $3\bm{\sigma}_p$ (\ref{sigma3}) for a scenario assigned to index $q=4$ and the ground truth trajectory (\ref{gt}).}
\label{fig:confidencePrediction}
\vspace{-8pt}
\end{figure}
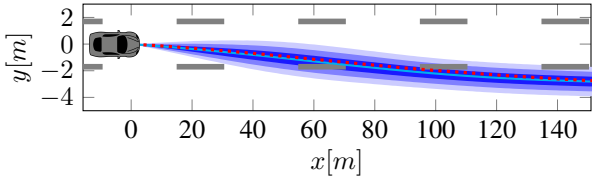
\subsection{Prediction performance}
To ensure fair benchmarking, all architectures are trained and tested on the same data $\mathcal{D}_\textrm{train}$ and $\mathcal{D}_\textrm{test}$. \mbox{Tab. \ref{tab:comparison}} compares the prediction accuracy of the proposed cVMD and state-of-the-art approaches based on the metrics ADE and FDE. For cVMD, only one trajectory prediction per scenario was generated and evaluated. Although the proposed cVMD did not outperform the best-performing prediction model, GFTNNv2, it demonstrated superior predictive capabilities to the other leading models in the field. However, this outcome was somewhat anticipated due to the fundamental differences between the proposed model and the existing ones. Unlike the models being compared, the proposed diffusion-based cVMD considers the inherent uncertainties related to the future trajectories of traffic participants. DDPMs rely on stochastic processes to generate future trajectories. 
While this stochasticity generally allows the model to produce multiple plausible predictions, it can also cause the model to have inferior performance compared to deterministic models. Nevertheless, the inherent stochasticity of DDPMs can be used to its advantage. DDPMs allow for the generation of a set of potential trajectories, derived from the same initial condition. This spectrum of possible trajectories can be used to approximate a statistical confidence interval, representing the range within which the actual trajectory is likely to fall a certain percentage of the time. Fig.~\ref{fig:confidencePrediction} illustrates this concept. As an example, eight generated trajectories for a scenario assigned to index $q=4$ are converted to a mean trajectory prediction $\bm{\mu}_p$ with a confidence interval constrained by the standard deviation $\bm{\sigma}_p$. Note that the observed variance within the generated trajectories is related to the parameterization of guidance scale $w$, linking the model uncertainty $\delta_m$ within the latent space to the confidence interval of the trajectory prediction (cf. Eq.~\ref{eq:adapw}). Such a confidence interval is an effective way to represent the uncertainty associated with these predictions and provides a measure of the reliability. 
\section{Conclusion}
\label{sec:conclusion}
The proposed cVMD architecture for vehicle trajectory prediction in interactive highway scenarios allows the generation of guaranteed drivable trajectories while taking into account the inherent multimodality of real-world scenarios. Unlike fully data-driven prediction methods, cVMD includes non-holonomic motion constraints and physical limitations into the generative prediction module. Another unique feature of cVMD is its ability to quantify the model's prediction uncertainty. Incorporating model uncertainty into the trajectory prediction process has been shown to improve the network's trajectory prediction performance. When evaluated on the publicly available highD dataset, cVMD demonstrated highly competitive capabilities with established state-of-the-art architectures.\\ 
A notable limitation of the diffusion-based cVMD is its extended inference time, which currently prevents it from being used in real-time applications. Minimizing the cVMD's inference time will be the focus of future efforts. Furthermore, experiments have shown that there are limits to the effectiveness of the context conditioning module used, which should be improved with further research.\\
\textbf{Acknowledgement.} The work was supported by \mbox{Audi AG}.
{
    \small
    \bibliographystyle{ieeenat_fullname}
    \bibliography{ref/ref_vmd}
}


\end{document}